\documentclass[sigconf]{acmart}
\usepackage{balance}
\AtBeginDocument{%
  \providecommand\BibTeX{{%
    \normalfont B\kern-0.5em{\scshape i\kern-0.25em b}\kern-0.8em\TeX}}}

\copyrightyear{2021}
\acmYear{2021}
\setcopyright{rightsretained}
\acmConference[MM '21]{Proceedings of the 29th ACM International Conference on Multimedia}{October 20--24, 2021}{Virtual Event, China}
\acmBooktitle{Proceedings of the 29th ACM International Conference on Multimedia (MM '21), October 20--24, 2021, Virtual Event, China}\acmDOI{10.1145/3474085.3479207}
\begin{document}
\fancyhead{}

\title{CLIP4Caption: CLIP for Video Caption}

\author{Mingkang Tang$^{1,2}$, Zhanyu Wang$^{1}$,Zhenhua Liu$^1$, Fengyun Rao$^1$, Dian Li$^1$, Xiu Li$^2$}
\affiliation{$^1$Kandian Content AI Lab, Platform and Content Group, Tencent\\
        $^2$Shenzhen International Graduate School, Tsinghua University, China}
\email{{mktang, zhanyuwang, edinliu, fengyunrao, goodli}@tencent.com, li.xiu@sz.tsinghua.edu.cn}

\renewcommand{\shortauthors}{Wang and Tang, et al.}

\begin{abstract}
Video captioning is a challenging task since it requires generating sentences describing various diverse and complex videos. Existing video captioning models lack adequate visual representation due to the neglect of the existence of gaps between videos and texts. To bridge this gap, in this paper, we propose a CLIP4Caption framework that improves video captioning based on a CLIP-enhanced video-text matching network (VTM). This framework is taking full advantage of the information from both vision and language and enforcing the model to learn strongly text-correlated video features for text generation. Besides, unlike most existing models using LSTM or GRU as the sentence decoder, we adopt a Transformer structured decoder network to effectively learn the long-range visual and language dependency. Additionally, we introduce a novel ensemble strategy for captioning tasks. Experimental results demonstrate the effectiveness of our method on two datasets: 1) on MSR-VTT dataset, our method achieved a new state-of-the-art result with a significant gain of up to 10\% in CIDEr; 2) on the private test data, our method ranking 2nd place in the ACM MM multimedia grand challenge 2021: Pre-training for Video Understanding Challenge. It is noted that our model is only trained on the MSR-VTT dataset.
\end{abstract}

\ccsdesc[500]{Computing methodologies~Neural networks}

\keywords{video caption, video-text matching, pre-train, transformer}

\maketitle

\section{Introduction}
Describing video content is a labor-intensive task for humans. Therefore, computer scientists have put much effort into connecting human language with visual information to develop a system that automatically describes videos using natural language sentences. The advancement of video captioning enhances various applications in reality, e.g., automatic video subtitling, aid to visually impaired person, human-computer interaction, and improving online video search or retrieval~\cite{amirian2020automatic}. 

    Early research in video captioning used template-based methods~\cite{guadarrama2013youtube2text,rohrbach2013translating,xu2015jointly}, which aligns predicted words with the pre-defined template. S2VT ~\cite{venugopalan2015sequence} proposed a LSTM~\cite{1997Long} based sequence-to-sequence model for video captioning. Since then, numerous sequence learning methods, which adopt encoder-decoder architecture to generate caption flexibly, were introduced~\cite{pan2016jointly,yao2015describing,yu2016video,pan2017video,wang2018reconstruction,Li_2018_CVPR}. ~\cite{yao2015describing} propose an attention-based approach that takes into account both the local and global temporal structure of videos to product descriptions. RecNet~\cite{wang2018reconstruction} proposed a reconstruction network that leverages both the video-to-text and text-to-sentence flows for video captioning. In recent years' study, some researchers also successfully use vision-language (VL) pretraining for VL understanding, which has made significant progress in the downstream task of image captioning~\cite{li2020oscar,zhou2020unified,li2021scheduled}.
    
All the methods mentioned above build their video encoder with a CNN-based network, lacking adequate visual representation since they only take advantage of the information from vision modality. In this paper, we introduce a video-text matching network which empowered by a well-pretrained CLIP~\cite{radford2021clip} model to learn the video embeddings taking fully advantage from both vision and language modality.

    Specially, we first pre-train a video-text matching model to obtain a text-correlated video embeddings, and then we taken those enhanced video embedding as input to fine-tune in a well pre-trained transformer decoder network. It is noted that our transformer decoder is initialized by the part of weights of the pretrained Uni-VL~\cite{luo2020univl} model. Extensive experiments demonstrate that our methodology outperforms state-of-the-art video captioning methods~\cite{sun2019multimodal} on the MSR-VTT dataset~\cite{xu2016msr}. Additionally, our methodology ranks 2nd in the ACM MM grand challenge 2021: Pre-training for Video Understanding Challenge, in the first track of pre-training for video captioning.
    
    
    The main contributions of this work are summarized as follows:
     
    $\bullet$ We utilize a CLIP-enhanced video-text matching network to enforce our model to learn strongly correlated video and text features for text generation.
     
    $\bullet$ We leverage the weights of well pre-trained video and language model Uni-VL while greatly simplified its structure to better-fitting video captioning tasks.

    $\bullet$ We design a novel ensemble mechanism for video captioning.
    
    $\bullet$ We extensively validate our model on the most widely used MSR-VTT dataset. The results indicate that our framework outperforms multiple state-of-the-art methods in video captioning, exhibiting the great potential of our framework for this challenging task.

\section{METHODOLOGY}
\begin{figure*}
  \centering
  \includegraphics[width=\textwidth]{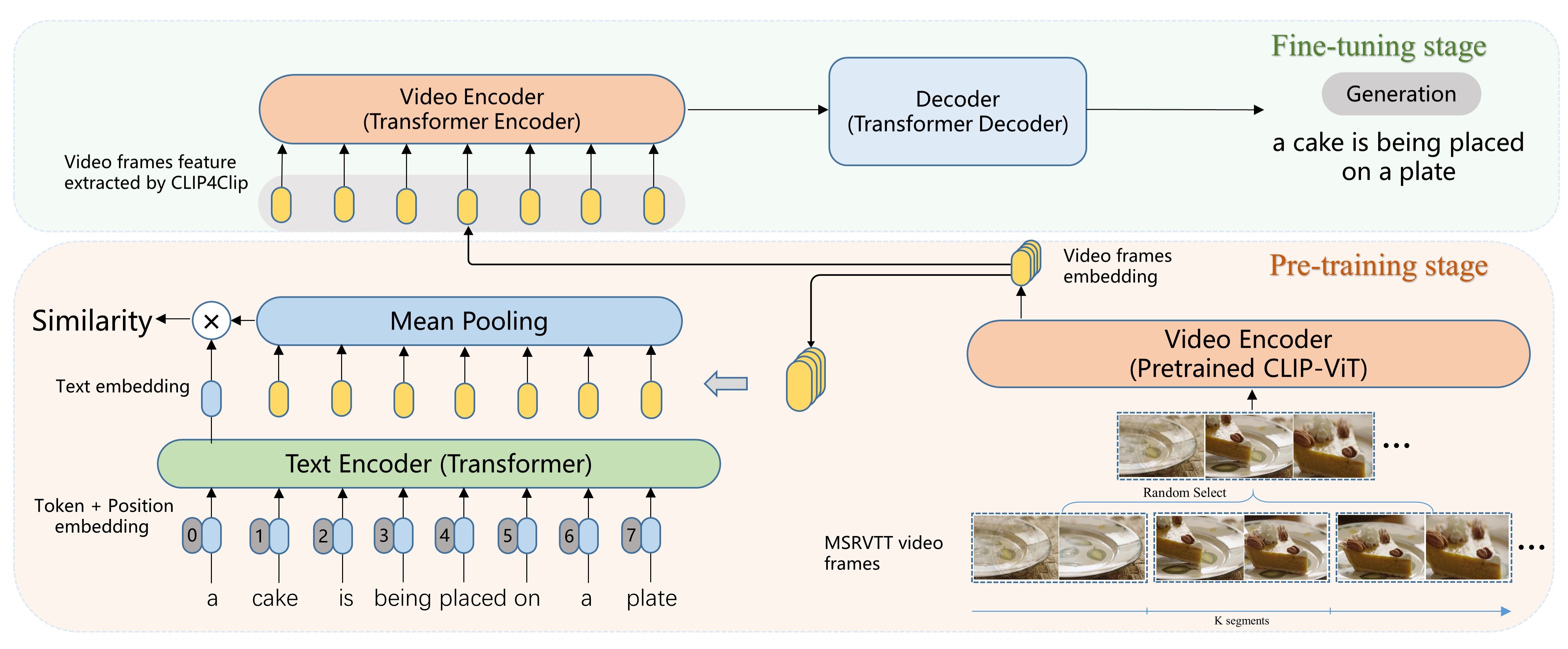}
  \caption{An Overview of our proposed CLIP4Caption framework comprises two training stages: a video-text matching pre-training stage and a video caption fine-tuning stage. In the pre-training stage, we build our video-text matching network upon a CLIP-based Video Encoder and a BERT-based Text Encoder, and it compares the similarity of video and text features for the match. In the fine-tuning stage, we take the strongly text-correlated video feature as input and fine-tuning our transformer decoder network for captioning task.}
  \Description{CLIP4Caption}
  \label{fig1}
\end{figure*}
Figure \ref{fig1} illustrates the framework of our proposed CLIP4Caption for video captioning. We train our system in two stage. First we pre-train a video-text matching network on MSR-VTT dataset to obtain better visual representation(\ref{section1}) (lower part in Fig.~\ref{fig1}). Second, we take our pre-trained matching network as the video feature extractor in fine-tuning stage (upper part in FIg.\ref{fig1}). A sequence of frames embedding was inputted to video encoder, connected with a decoder which generated the text (\ref{section2}). For ensemble, we train multiple caption models with different layers of encoder and decoder, and ensemble all the generated captioning text for a final strong result (\ref{section3}). Details will be elaborated as follows.

\subsection{Video-text matching pre-training}\label{section1}
As the CLIP4Clip model transferred from CLIP~\cite{radford2021clip} has demonstrated outstanding performance in the video-text retrieval task, we pre-train our video-text matching network (VTM) upon CLIP4Clip. CLIP4Clip extracts frames of images from the video at 1 FPS, the input video frames for each epoch come from the video's fixed position. We improve the frames sampling method to the TSN sampling\cite{wang2016temporal}, which divides the video into K splits and randomly samples one frame in each split, thus increasing the sample randomness on the limited data set. After the TSN sampling, input frames are encoded by the pre-trained CLIP (ViT-B/32) video encoder~\cite{dosovitskiy2020image} with 12 layers and the patch size 32, since CLIP is adequate for the image-text retrieval task. Take the input video frames of video as $v_{i}=\{v_i^1,v_i^2,\dots,v_i^{|v_i|}\}$, the video frames embedding can denote as $f_{i}=\{f_i^1,f_i^2,\dots,f_i^{|v_i|}\}$.
    
    A 12-layer 512-wide model with eight attention heads Transformer~\cite{vaswani2017attention} encoder is used as text encoder, whose weights originated from the pre-trained CLIP text encoder. Following CLIP and CLIP4Clip, the [EOS] token’s activations of the highest layer of the transformer are used as the feature representation of the input text. For the input text $s_{j}$, corresponding text embedding is denoted as $t_{j}$.
    
    After video encoding, we use a mean pooling layer to aggregate the embedding of all frames and obtain a average frame embedding. Then, the similarity function can be defined for video-text matching. Similar to CLIP4Clip, we adopt cosine similarity to measure similarity between joint video frames embedding $\hat{f}_{i}$ and text embedding $t_{j}$, as formulated by:

    \begin{equation}
        \label{eq:1}
        s(\hat{f}_{i},t_{j}) = \dfrac{{t_{j}}^\top\hat{f}_{i}}{\| {t_{j}}\|  \|\hat{f}_{i}\|}.
    \end{equation}
    
    Video-text matching is trained in a self-supervised way. Given a batch of N video-text pairs, VTM generates a N × N similarities and the optimization goal is to maximize the similarity between paired video-text and maximize the similarity of unpaired text. Therefore, the loss function is defined as follows:
    
    \begin{equation}
    \label{eq:2}
    \mathcal{L}_{v2t}=-\dfrac{1}{N} \sum_{i}^{N} \log\dfrac{\exp(s(\hat{f}_{i},t_{i}))}{\sum_{j}^{N}\exp(s(\hat{f}_{i},t_{j}))},
    \end{equation}
    \begin{equation}
    \label{eq:3}
    \mathcal{L}_{t2v}=-\dfrac{1}{N} \sum_{i}^{N} \log\dfrac{\exp(s(\hat{f}_{i},t_{i}))}{\sum_{j}^{N}\exp(s(\hat{f}_{j},t_{i}))},
    \end{equation}
    \begin{equation}
    \label{eq:4}
    \mathcal{L}=\mathcal{L}_{v2t}+\mathcal{L}_{t2v},
    \end{equation}
    where $\mathcal{L}_{v2t}$ and $\mathcal{L}_{t2v}$ denotes the loss function of video-to-text and text-to-video respectively.
    
    We use the output of ViT video encoder, a sequence of frames embedding, as our video visual representation. Each frame is mapped to 512d visual feature, resulting $n \times 512$ dynamic features $F_{v_{i}} = ViT(v_{i})$ for each video, where n is the number of frames.
\subsection{Fine-tune on video captioning}\label{section2}
In fine-tuning stage, we leverage the well pre-trained model Uni-VL and fine-tune the encoder-decoder architecture of Uni-VL on video captioning with MSR-VTT dataset. Uni-VL is a two-stream video and language pre-training model. During Uni-VL's pre-training, text and video are input to text encoder and video encoder, respectively, and a cross encoder aligns the text embedding and video embedding. In our fine-tuning stage, we discard the text encoder and cross encoder since the input video of our dataset without related transcripts. The fine-tuning stage on total pre-trained Uni-VL layers is difficult since the MSR-VTT dataset (for fine-tuning stage) is relatively tiny to the Uni-VL's pre-training dataset HowTo100M~\cite{miech2019howto100m}. CLIP4Caption, therefore, train effortless and prevent over-fitting through reducing the number of Transformer layers. 

    As described above, our captioning model is composed of the Transformer based Video Encoder and Decoder, the strongly text-correlated video feature $F_{v}$ is input to one-layer Transformer Video Encoder(TE) to obtain the enhanced feature $f_{ve} = TE(F_{v})$, and then fed into a three-layer Transformer Decoder (TD) to produce caption $t = TD(F_{ve})$ for each video. We initialize TE and TD with the weights pre-trained in Uni-VL and train the model with only cross-entropy loss:
    \begin{equation}
    \label{eq:5}
    \mathcal{L}_{ce}=-\sum_{t=1}^{L}\log{p_{t}(S_{t})},
    \end{equation}
    where L is the max length of caption sentence, $p_{t}$ is the probability of predicted word at time t, and $S_{t}$ is the sentence which has generated at time t.
\subsection{Ensemble strategy}\label{section3}
The single model is not strong enough for a great predicted result. In order to obtain a more powerful caption result, we design a novel metric-based voting strategy for captioning task. We utilise the captioning evaluation metrics, such as BLEU4, CIDEr, SPICE, etc., as the ``importance score” of a generated sentences and select the sentence with highest score to compose the final result. 

    Mathematically, Considering the predicted captions of one video from $n$ different models as $T_{i}$, the importance score for $ith$ caption $S_{i}$ using single metric can be calculated by assuming the rest of predicted captions $[T_{j}]_{j \neq i}$ as "ground-truth" captions:
    \begin{equation}
    \label{eq:6}
    S_{i} = metric(ref=\left[T_{1},\dots,T_{i-1},T_{i+1}, \dots, T_{n}\right], hpy=T_{i}),
    \end{equation}
    where $i \in \left[1, n\right]$ and $metric(\cdot)$ is the captioning metric. The predicted caption with biggest score $S$ is selected as the final output. Since captioning task often uses multiple metrics, and the value range of each metric is inconsistent, we use the maximum value of each metric to normalize it~\cite{chen2020semantics}. Considering multiple metrics, the overall metric can be calculated as:
    \begin{equation}
    \label{eq:7}
    metric_{overall} = (\dfrac{metric_{1}}{metric_{1b}}+\dfrac{metric_{2}}{metric_{2b}}+\dots+\dfrac{metric_{M}}{metric_{Mb}})/M,
    \end{equation}
    where M denotes the number of metrics used for calculating $metric_{overall}$, $metric_{ib}$ denotes the best numeric value of the specific metric $metric_{i}$.
    And the importance score of multiple metrics is:
    \begin{small}
    \begin{equation}
    \label{eq:8}
    \hat{S}_{i} =
    metric_{overall}(ref=\left[T_{1},\dots,T_{i-1},T_{i+1}, \dots, T_{n}\right], hpy=T_{i}),
    \end{equation}
    \end{small}
\section{RESULTS}
\subsection{Dataset split}
On account of the significant difference between Video Understanding Challenge pre-training dataset Auto-captions on GIF (ACTION) \cite{pan2020auto} and the MSR-VTT data set, we only use MSR-VTT as our training dataset. The MSR-VTT dataset covers a broad range of video content categories, with a total video time of about 50 hours, 10000 videos, and 20 descriptions per video. For pre-training results in Table ~\ref{table1} and Table ~\ref{table2}, we report our results on the split of MSR-VTT Training-9K\cite{luo2021clip4clip}, which was used in CLIP4Clip. For fine-tuning results in Table ~\ref{table3}
, we used the MSR-VTT’s standard split (MSR-VTT Training-6K), i.e., 6,512, 498, and 2,990 clips for training, validation, and testing for comparision with state-of-the-art methods. We also used total MSRVTT (MSR-VTT Training-10K) to train more models for our ensemble results in Table ~\ref{table4}, and evalute our ensemble mechanism on Video Understanding Challenge test set.
\subsection{Pre-training result}
\begin{table}
\caption{Pre-training results of video-text matching on MSR-VTT Training-9K. In the column $R_{T2V}$@K denotes text-to-video recall at rank K. For VTM, max frames K means K splits for TSN sampling. }
\Description{VTM}
\setlength{\tabcolsep}{1.5mm}{
\begin{tabular}{lcccc}
\hline
Methods       & max frames & $R_{T2V}$@1 & $R_{T2V}$@5 & $R_{T2V}$@10 \\ \hline
CLIP4Clip     & 12 & 42.6   & 70.1   & 80.2 \\
VTM & 12 & 42.7   & 70.1   & \textbf{81.2}   \\
VTM & 6  & 44     & 70.6   & 80    \\
VTM & 3  & \textbf{44.5}     & \textbf{71.1}   & 79.7    \\ \hline
\end{tabular}}
\label{table1}
\end{table}

\begin{table}
\centering
\caption{Pre-training results of video-text matching on MSR-VTT Training-9K. In the column $R_{V2T}$@K denote video-to-text recall at rank K. For VTM, max frames K means K splits for TSN sampling. }
\Description{CLIP4Clip}
\setlength{\tabcolsep}{1.5mm}{
\begin{tabular}{lcccc}
\hline
Methods             & max frames & $R_{V2T}$@1 & $R_{V2T}$@5 & $R_{V2T}$@10 \\
\hline
CLIP4Clip     & 12 & 43.4   & 70.2   & 81.7 \\
VTM   & 12 & 43.4   & 70.5   & \textbf{82.2}   \\
VTM & 6 & 43.5     & 71.1   & 81.4    \\
VTM & 3 & \textbf{44.3}     & \textbf{71.7}   & 81.7    \\
\hline
\end{tabular}}
\label{table2}
\end{table}


Video-text matching pre-training is done on 8 NVIDIA Tesla P40 GPU graphics, with batch size set to 512 and max epochs set to 5. We use standard retrieval metrics: recall at rank K, denoted as R@K, to evaluate the performance of our pre-trained network. We report the metrics of R@1, R@5, R@10 for both text-to-video retrieval and video-to-text retrieval. The VTM result with different TSN split is shown in Table ~\ref{table1} and Table ~\ref{table2}. Our VTM outperformed origin CLIP4Clip, and TSN sampling with 3 splits perform better in R@1 and R@5. Given this, we used VTM with 3 splits TSN as our feature extractor, and retrained the VTM in MSR-VTT Training-6K in the fine-tuning stage for better visual representation.
\subsection{Fine-tuning result}
\begin{table}
\centering
\caption{Comparison on MSR-VTT dataset. ``B@”, ``R”, ``M”, ``C”, ``S” denotes ``BLEU-4”, ``ROUGE-L”, ``METEOR”, ``CIDEr”, ``SPICE” respectively. $\dagger$ indicates the results are quoted from the published literature~\cite{perez2021bridging}. For other methods in comparison, their results are obtained by re-running the publicly released codes or models on MSR-VTT dataset using the same training-test partition as our method.}
\begin{tabular}{lccccc} 
\hline
Methods & B@4   & R     & M     & C     & S      \\
\hline
$REVnetv3-RL^{\dagger}$~\cite{li2019revnet} & 42.4 & 62.3 & 28.1 & 53.2 & - \\
$ORG-TRL^{\dagger}$~\cite{zhang2020object} & 43.6 & 62.1 & 28.8 & 50.9 & - \\
$CST\_GT\_None^{\dagger}$~\cite{phan2017consensus} & 44.1 & 62.4 & 29.1 & 49.7 & - \\
$SCN-LSTM+sampling^{\dagger}$~\cite{chen2020semantics} & 43.8 & 62.4 & 28.9 & 51.4 \\
$topic-guided^{\dagger}$~\cite{chen2019generating} & 44.9 & 62.8 & 29.6 & 51.8 &- \\
$GFN-POS\_RL(IR+M)^{\dagger}$~\cite{wang2019controllable}          & 41.3 & 62.1 & 28.7 & 53.4 & -      \\
$VNS-GRU^{\dagger}$~\cite{chen2020delving} & 46.0 & 63.3 & 29.5 & 52.0 & - \\
$MSAN_{f+o+c}^{\dagger}$~\cite{sun2019multimodal} & 46.8 & - & 29.5 & 52.4 & -      \\
$AVSSN$~\cite{perez2021attentive}                      & 42.8 & 61.7 & 28.8 & 46.9 & -      \\
$SemSynAN$~\cite{perez2021improving}                   & 44.3 & 62.5 & 28.8 & 50.1 & 6.3      \\
\hline
$Uni-VL$~\cite{luo2020univl}                     & 42.2 & 61.21   & 28.8 & 49.9 & 6.5  \\ 
$CLIP4Caption$ & 46.1 & 63.7 & 30.7 & 57.7 & 7.6  \\
$CLIP4Caption(ensemble)$ & \textbf{47.2} & \textbf{64.8} & \textbf{31.2} & \textbf{60.0} & \textbf{7.9}  \\
\hline  
\end{tabular}
\label{table3}
\end{table}

In the fine-tuning stage, we used the pre-trained VTM on MSR-VTT Training-6K as our feature extractor. To be consistent with Uni-VL, we set the maximum frame length to 20. Caption fine-tuning is done on 4 GPUs, making the batch size 1024 and the total epochs 30.

    In Table ~\ref{table3}, we report the standard captioning metrics, BLEU-4~\cite{papineni2002bleu}, ROUGE-L~\cite{lin2004rouge}, METEOR~\cite{banerjee2005meteor}, CIDEr~\cite{vedantam2015cider}, SPICE~\cite{anderson2016spice} of our proposed CLIP4Caption, and other state-of-the-art methods for video captioning on MSR-VTT dataset. As shown in Table \ref{table3}, our pre-training stage learns the powerful text-correlated visual representation for text generation, significantly improving all the metrics upon Uni-VL. CLIP4Caption achieved a new state-of-the-art result with a significant gains of up to 10\% in the CIDEr score.
\subsection{Ensemble result}
\begin{table}
\centering
\caption{Ensemble results of video captioning on the Video Understanding Challenge test set. "single model" means the best result using a single model, while "k models" indicate the ensemble strategy uses results from k models for a better result.}
\begin{tabular}{lcccc} 
\hline
Methods                    & B@4 & M & C & S  \\ 
\hline
CLIP4Caption(single model) & 22.76  & 17.89  & 26.93 & 5.93   \\
CLIP4Caption(9 models)     & 23.61  & 18.69  & 29.35 & 6.67   \\
CLIP4Caption(17 models)    & \textbf{23.78}  & 19.18  & 30.39 & 7.14   \\
CLIP4Caption(47 models)    & 23.42  & 19.40  & 30.56 & 7.47   \\
CLIP4Caption(59 models)    & 23.67  & \textbf{19.63}  & \textbf{31.19} & \textbf{7.53}   \\
\hline
\end{tabular}
\label{table4}
\end{table}
We vary dataset split and the layers of the Transformer to train more models. We used two splits for the pre-training stage, the standard dataset split MSR-VTT Training-6K, and the other dataset split MSR-VTT Training-9K, with 9000 videos for training and 1000 videos for validation. As a result, two VTM features are obtained: VTM-6K-feature and VTM-9K-feature. 

    We adopted three 
    splits for the fine-tuning stage, MSR-VTT Training-6K and MSR-VTT Training-9K as used in the pre-training stage, and  MSR-VTT Training-10k, which used the total MSR-VTT dataset. At the same time, we also combined the different layers of Transformer layers. For the visual encoder, we tried 1-layer, 3-layer, and 6-layer Transformer encoders. As for the decoder, we tried the number of Transformer layers from 1 to 6. These combinations produced a lot of captioning results. We eliminated some of the results that were not effective on the MSR-VTT validation dataset and applied ensemble strategy in other results.

    We Validate our proposed strategy in Video Understanding Challenging test dataset. The metric used for the ensemble is SPICE and BLEU-4, because of the poor performance using other metrics. The experimental results are shown in Table \ref{table3}. Compared with the best result of a single model, our ensemble strategy has significantly improved the metrics on the test dataset, and ranks 2nd in the Video Understanding Challenge. Furthermore, the result is positively correlated with the number of results we use, which means we can use this ensemble strategy to improve our results by continuously training enough models.
\section{CONCLUSION}

In this work, we focus on learning better visual representation for text generation and improving video captioning with video and language pre-training models. We propose the CLIP4Caption, a two-stage language and video pre-training-based video caption solution. For better visual representation, we adopt the pre-training stage to learn strongly text-correlated video features. Also, to improve video captioning, we make use of Uni-VL pre-trained weights to initialize our encoder-decoder-based captioning architecture and fine-tune the model in MSR-VTT dataset. Besides, we introduce a novel ensemble strategy to ensemble multiple models' captioning results by using captioning metrics. Extensive experiments indicate that our proposed CLIP4Caption significantly outperforms the current state-of-the-art method and ranks 2nd in the Video Understanding Challenge test dataset.
\section{Acknowledgments}
This research was partly supported by the National Natural Science Foundation of China (Grant No. 41876098), and Shenzhen Science and Technology Project (Grant No. JCYJ20200109143041798).
\bibliographystyle{ACM-Reference-Format}
\balance
\bibliography{main}

\end{document}